\documentclass[conference]{IEEEtran}
\IEEEoverridecommandlockouts
\usepackage{cite}
\usepackage{amsmath,amssymb,amsfonts}
\usepackage{algorithmic}
\usepackage{graphicx}
\usepackage{textcomp}
\usepackage{xcolor}
\usepackage{mathtools}
\usepackage{soul}
\usepackage[T1]{fontenc}
\usepackage[hidelinks]{hyperref}

\newcommand{\onlineassessment}{DNN-OS}
\newcommand{\offlineassessment}{DeepEST}

\def\BibTeX{{\rm B\kern-.05em{\sc i\kern-.025em b}\kern-.08em
    T\kern-.1667em\lower.7ex\hbox{E}\kern-.125emX}}
\begin{document}

\title{Iterative Assessment and Improvement\\of DNN Operational Accuracy}

\author{\IEEEauthorblockN{Antonio Guerriero, Roberto Pietrantuono, Stefano Russo}
\vspace{3pt}
\IEEEauthorblockA{\textit{DIETI, Universit\`a degli Studi di Napoli Federico II}\\
Via Claudio 21, 80125 - Napoli, Italy \\
\{antonio.guerriero, roberto.pietrantuono, stefano.russo\}@unina.it}
}

\maketitle

\begin{abstract}
Deep Neural Networks (DNN) are nowadays largely adopted in many application domains 
thanks to their human-like, or even superhuman, performance in specific tasks.
However, due to unpredictable/unconsidered operating conditions, unexpected failures show up on field,
making the performance of a DNN in operation very different from the one estimated prior to release. 

In the life cycle of DNN systems, the assessment of accuracy is typically addressed in two ways: offline, via sampling of operational inputs, or online, via pseudo-oracles.
The former is considered more expensive due to the need for manual labeling of the sampled inputs. The latter is automatic but less accurate. 

We believe that emerging iterative industrial-strength life cycle models for Machine Learning systems, like MLOps, offer the possibility to leverage inputs observed in operation not only to provide faithful estimates of a DNN accuracy, but also to improve it through remodeling/retraining actions.

We propose DAIC (DNN Assessment and Improvement Cycle), an approach which combines “low-cost” online pseudo-oracles and “high-cost” offline sampling techniques to estimate and improve the operational accuracy of a DNN in the iterations of its life cycle. 
Preliminary results show the benefits of combining the two approaches and integrating them in the DNN life cycle.
\end{abstract}


\section{Introduction}

Nowadays, Machine Learning (ML) finds large adoption in various application domains. This trend is due to the ability of ML, in particular of Deep Neural Networks (DNN), to reach human beings' effectiveness in many tasks \cite{Kuhl20, He15, Silver17}.

The reliability of ML systems is usually measured in terms of \textit{accuracy}. In the case of classification, the accuracy is computed as the number of correctly classified examples out of the total. 
The difficulty to automate the assessment of DNN accuracy still represents a threat to their application also in critical domains.

The main activities related to evaluating the accuracy and consequently improving the DNN are typically executed before its release in the execution environment. Metamorphic testing \cite{Xie2018} and mutation testing \cite{Ma2018, Li22} represent the most common strategies to evaluate the robustness of the DNN and to forecast the reliability of these systems in the operational environment.

However, the accuracy estimated before release can substantially diverge from the one obtained during  operation (\textit{operational accuracy}). Retch \textit{et al.} demonstrated how the accuracy scores of classifiers can significantly drop when completely new data are submitted \cite{Recht19}.
This problem grows up when unexpected phenomena occur in operation, such as \textit{distribution shift} or \textit{label shift} \cite{Garg20}. 

Iterative life cycles specific for DNN - such as MLOps \cite{Alla21, GoogleMLOps} - have been envisaged by companies like Google. In these DNN life cycle models, development and operational stages are linked in a loop \cite{Ashmore2021}, aiming to assess and improve the accuracy the DNN according to the operating conditions. In particular, they may exploit operational data for remodeling and/or retraining the DNN before the new deployment (\textit{experimental stage}), and for both the automatic evaluation of the accuracy and the automatic re-training of the models on the field (\textit{deployment stage}).

The true label of operational data collected by monitoring the DNN is generally unknown. 
This is a general issue in software testing, known as the \textit{oracle problem} \cite{Barr15}. For ML systems, according to Murphy \textit{et al.}, “\textit{there is no reliable test oracle to indicate what the correct output should be for arbitrary input}” \cite{Murp2008}. The problem clearly affects also DNN accuracy estimation \cite{Guerriero20}. %

Two approaches to assess DNN operational accuracy are: 
\begin{itemize}
\item[i)] to automatically evaluate the correct classification of operational inputs by means of \textit{pseudo-oracles} (often in turn based on ML models), which may detect mispredictions based on various sources of knowledge; %
\item[ii)] to reduce the size of the operational dataset to be labelled, by proper statistical sampling of few representative inputs. %
\end{itemize}

Pseudo-oracles do not need human intervention; however, they typically suffer from a high number of false positives \cite{Zhang20}, due to the probabilistic nature of the knowledge used to evaluate the output of the DNN under assessment.
Sampling techniques may reduce but not avoid costly and time-consuming feedback from a human oracle; however, they avoid false positives, and provide more faithful estimates of the DNN operational accuracy. %

This paper proposes the DNN Assessment and Improvement Cycle (DAIC), integrating automatic assessment via pseudo-oracles and assessment via sampling. %
Its objectives are to provide faithful estimates of the operational accuracy of a DNN while reducing the cost of manual intervention, and to exploit the new labeled examples to take remodeling/retraining actions to improve the DNN accuracy.

The preliminary results of experiments with the MNIST handwritten digits dataset \cite{LeCun2010} show that DAIC is effective in providing DNN accuracy estimates, by leveraging automatic pseudo-oracles to follow the accuracy of the DNN with unlabeled samples, %
and triggering the high-cost sampling-based assessment only when necessary to update estimates. 
Collected operational samples are then further exploited to improve the accuracy of the DNN in an iteration cycle. 
DAIC is robust to phenomena like label shift.

The paper is structured as follows. Section \ref{sec:OAA} describes the techniques for the operational accuracy assessment of DNN. Section \ref{sec:aic} introduces the DNN Assessment and Improvement Cycle; Section \ref{sec:prel} presents the preliminary results. Section \ref{sec:future} describes future plans; Section \ref{sec:conclusions} presents the conclusions.

\section{Operational accuracy assessment of DNN}
\label{sec:OAA}
\subsection{Assessment via pseudo-oracles}
Automatic pseudo-oracles are typically built using \textit{cross referencing} \cite{Srisak18,Pei19,Wang20} based on the knowledge encoded into the training set. This knowledge is extracted through multiple implementations -- diverse from each other (e.g., different ML models, or same ML model but different architectures) - %
to perform a majority voting. 
These techniques are strictly affected by biases in the training set. When training data are not representative of the operational environment, performance of that oracles degrades significantly.

Other techniques have been proposed to extract knowledge from the training data to build automatic oracles, for instance, by using dedicated networks (ConfidNet \cite{Corbiere19} and autoencoders \cite{Stocco2020}) or exploiting features of the system under assessment itself (e.g. the output of internal layers \cite{Xiao21}).

Techniques considering only the training dataset and the DNN as knowledge to build a pseudo-oracle are particularly sensitive to deviations of the operational context from the pre-deployment one.
Therefore, they are expected to poorly perform in presence of phenomena like label shift \cite{Garg20,Candela09}. Supervised DNN algorithms face a label shift when the distribution of the labels of inputs changes with respect to training, despite everything else remains unchanged: in practice, when unlabeled operational inputs are similar to training examples, thus are classified by the DNN as per training, yet their actual class is different from the one learnt during training.

For image classification problems, the ICOS oracle surrogate has been proposed to assess the accuracy of Convolutional Neural Networks after their release in operation \cite{Guerriero23}. ICOS extracts invariants form different sources of knowledge to evaluate unlabeled operational examples.

Similarly to ICOS, we consider a pseudo-oracle -- hereafter called \onlineassessment{} - which exploits three different sources of knowledge (the operational domain, training data, and the DNN) to define three set of invariants (\textit{domain}, \textit{data}, \textit{model}), used to automatically evaluate the output of the DNN under assessment. 

An example of \textit{domain invariant} for an autonomous driving vehicle, assuming a street with a speed limit of $50$ $km/h$, is: $fail \textrm{ :- } speed\_limit =  50$ $km/h, accelerate = true, current\_speed = 50$ $km/h$.
Such domain invariants allow the oracle to detect failures looking at the output of the DNN and at its effect on the whole system.
The usage of domain invariants makes \onlineassessment{} robust against unexpected phenomena in operation with respect to the state-of-the-art techniques.

\textit{Data} and \textit{model invariants} can be automatically extracted from the training and validation datasets with a ML algorithm. These invariants look for patterns in the input data and the DNN, respectively, such as a subset of pixels (for \textit{data}) or neurons (for the \textit{model}) that always assume specific values when a failure occurs. 

For its characteristics, the assessment via pseudo-oracle can be performed \textit{online}, namely when the system is in operation. The automated oracle computes the accuracy on %
actual inputs.
This estimate can be used to suggest to the testers if correcting/improving actions are needed.

The online assessment %
is characterized by a fixed cost for the “knowledge extraction” and parameters tuning of the pseudo-oracle  algorithm, which occur \textit{una tantum}.

\subsection{Assessment via sampling}

The usage of sampling to reduce the cost of the manual labeling of operational examples has been explored in the recent literature \cite{Li19,Chen20,Gao22,Guerriero21}.
Some techniques are used to select a small data sample that accurately represents the population \cite{Li19,Chen20,Gao22} to obtain a faithful estimate of the accuracy provided during operation. A representative sample would roughly contain the same proportion of examples causing misprediction as the operational dataset.

However, the mere imitation of the expected input can be inefficient, especially with very accurate DNN, because of the great effort to manually label correctly classified examples to get an acceptable estimate of the operational accuracy. This cost makes it evident that maximizing the sampling of examples related to wrong outputs, while still getting an unbiased estimate of the operational accuracy, is preferable.

With \offlineassessment{}, Guerriero \textit{et al.} \cite{Guerriero21} aim to both provide faithful estimates of the operational accuracy, but trying to sample more failures examples (e.g. misclassifications) and balancing the unequal sampling during the estimation process. 

This assessment strategy can be performed \textit{offline}, namely when the monitored operational data are available together with the outputs of the DNN. With this data, an estimate of the accuracy can be computed via sampling and manual labeling. 
The high cost of manual labeling the operational input is balanced by the possibility to use the labeled examples to take improving actions for the DNN under assessment.

\section{DNN Assessment and Improvement Cycle}
\label{sec:aic}
A way to reduce the cost of applying and maximize the benefit is to combine the online and offline assessment in a cycle, called DNN Assessment and Improvement Cycle (DAIC). The idea is to have at each cycle a “low-cost” estimate of the accuracy provided through a pseudo-oracle, and to trigger a “high-cost” (but more faithful) offline sampling-based estimate only when the operational accuracy estimated by the automatic pseudo-oracle drops under a given threshold. 

\begin{figure*}[t]
	\centering
	 	\includegraphics[width=0.88\textwidth]{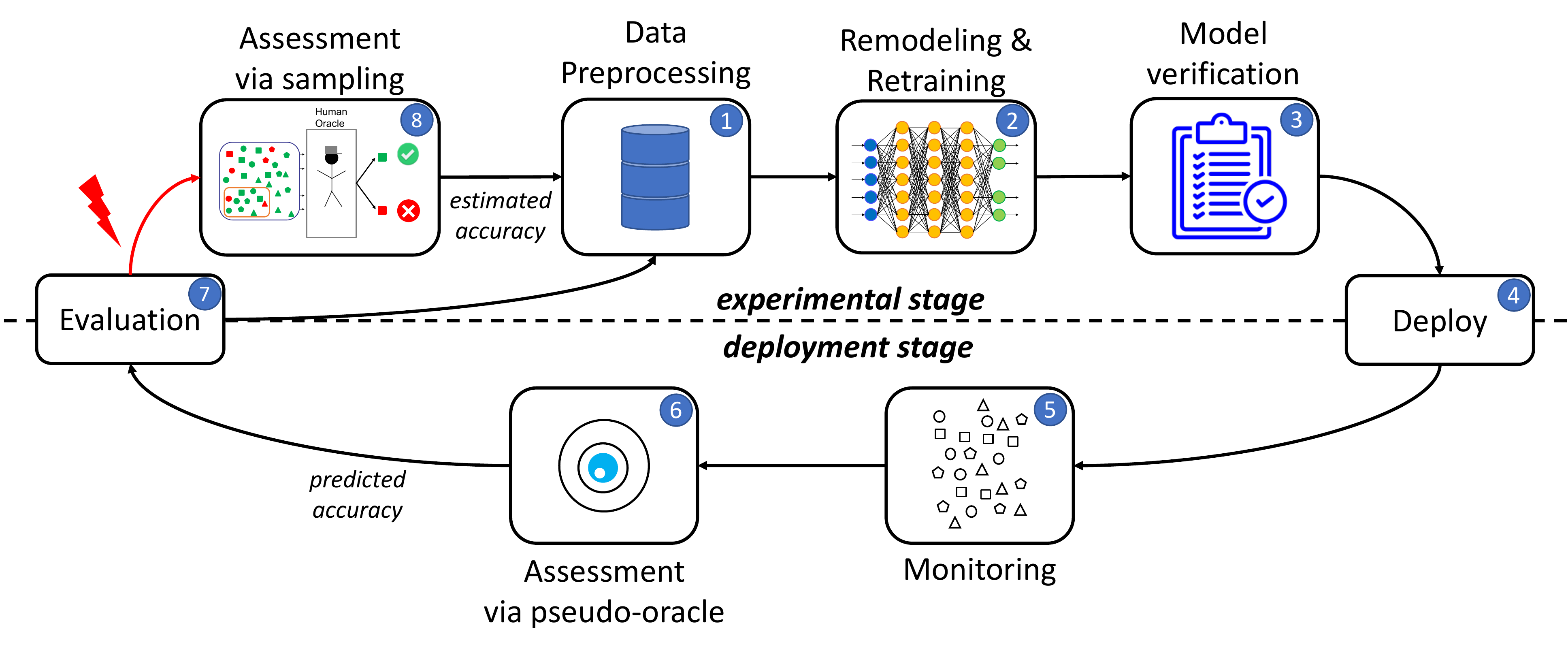}
	\caption{The proposed MLOps-like DNN Assessment and Improvement Cycle (DAIC)}
	\label{fig:accycle}
\end{figure*}

Like MLOps, DAIC entails an \textit{experimental} and a \textit{deployment stage}, with the following phases (Figure \ref{fig:accycle}):
\begin{enumerate}
    \item \textbf{Data Preprocessing}: in the starting phase of a cycle, the training and the verification datasets are updated considering new labeled examples (if available) and based on the accuracy estimate computed in the previous iteration. %
    This phase updates the training set so as to better represent the operating conditions actually observed.
	\item \textbf{Remodeling and Retraining}: the model is trained from scratch (first iteration or in case of re-modeling), or re-trained with the training set output of the Data Processing phase. 
	\item \textbf{Model verification}: the accuracy of the trained DNN is computed prior to release (\textit{verification accuracy}) on the verification dataset generated by the Data Preprocessing phase.
	\item \textbf{Deploy}: the DNN is deployed into the execution environment, and put in operation. %
	\item \textbf{Monitoring}: (unlabeled) inputs to the DNN and the corresponding DNN outcomes are collected; additional information on operating conditions (input sources, user typologies, operational profile, etc.) may be collected, if available, to build domain invariants.
	\item \textbf{Assessment via pseudo-oracle}: an automatic pseudo-oracle is used to classify each output of the DNN as \textit{Pass} or \textit{Fail}. The oracle predictions are used to compute an estimate of the DNN accuracy in operation, called \textit{predicted accuracy}.
	\item \textbf{Evaluation}: 	when the predicted and the verification accuracy diverge, an offline assessment session is triggered (8); otherwise, the sampling-based assessment is skipped.

	\item \textbf{Assessment via sampling}: a set of inputs is sampled and (manually) labeled, and an estimate of the operational accuracy is computed.
\end{enumerate}

The idea is to consider the pseudo-oracle for a continuous evaluation of the operational accuracy provided by the DNN to reduce the cost of manual labeling, retraining, and remodeling, performing them only when required.

\section{Experiments}
\label{sec:prel}

\subsection{Accuracy assessment algorithms, datasets, open artifacts}
DAIC experiments have been conducted with two pseudo-oracles and one sampling accuracy assessment algorithms.

The two pseudo-oracles are SelfChecker \cite{Xiao21} (the automatic oracle that exploits the features of the DNN under test itself to evaluate the predictions), and \onlineassessment{}. 

The sampling-based assessment algorithm is \offlineassessment{} \cite{Guerriero21}; it considers auxiliary variables, such as the \textit{confidence} of DNN predictions, to guide the sampling through as much as possible failing examples and to balance the unequal sampling in the estimation.

The dataset considered for the preliminary experiments is MNIST \cite{LeCun2010}, a famous dataset for handwritten digits classification. In particular, $1,000$ examples are considered for training, $500$ for the verification set, and $1,000$ unlabeled inputs (for each cycle) as the operational dataset. 

\onlineassessment{} invariants are obtained as follows:
\begin{itemize}
    \item \textit{Domain} invariants: defined by domain experts about the input sources of the DNN. In particular, for MNIST, we assume that users insert input into three different forms requiring respectively digits without straight lines \{0, 3, 6, 8, 9\}, digits with straight lines \{1, 4, 7\} only, and remaining digits \{2, 5\}. We define an invariant for each form. The output provided by the DNN for each operational input is checked against the set of possible digits expected for the source form.
    \item \textit{Data} invariants: automatically extracted from the training data in form of decision rules (\textit{C4.5} algorithm \cite{c45}) and filtered based on their confidence ($C \geq 0.99$) and support ($S \geq 10$).
    \item \textit{Model} invariants: extracted from validation data with \textit{Random Forest} using the output of the neurons of the last layer as features.
\end{itemize}
The sample size considered for \offlineassessment{} is $500$, and the proportion of examples sampled randomly with respect to those with weighted sampling is set to $0.5$. 

For independent verification or replication, the experimental code is available on GitHub at: 
\begin{center}
\url{https://github.com/dessertlab/DAIC.git}.
\end{center}
\vspace{-2pt}

\begin{figure}[t]
	\centering
 	\includegraphics[width=.99\columnwidth]{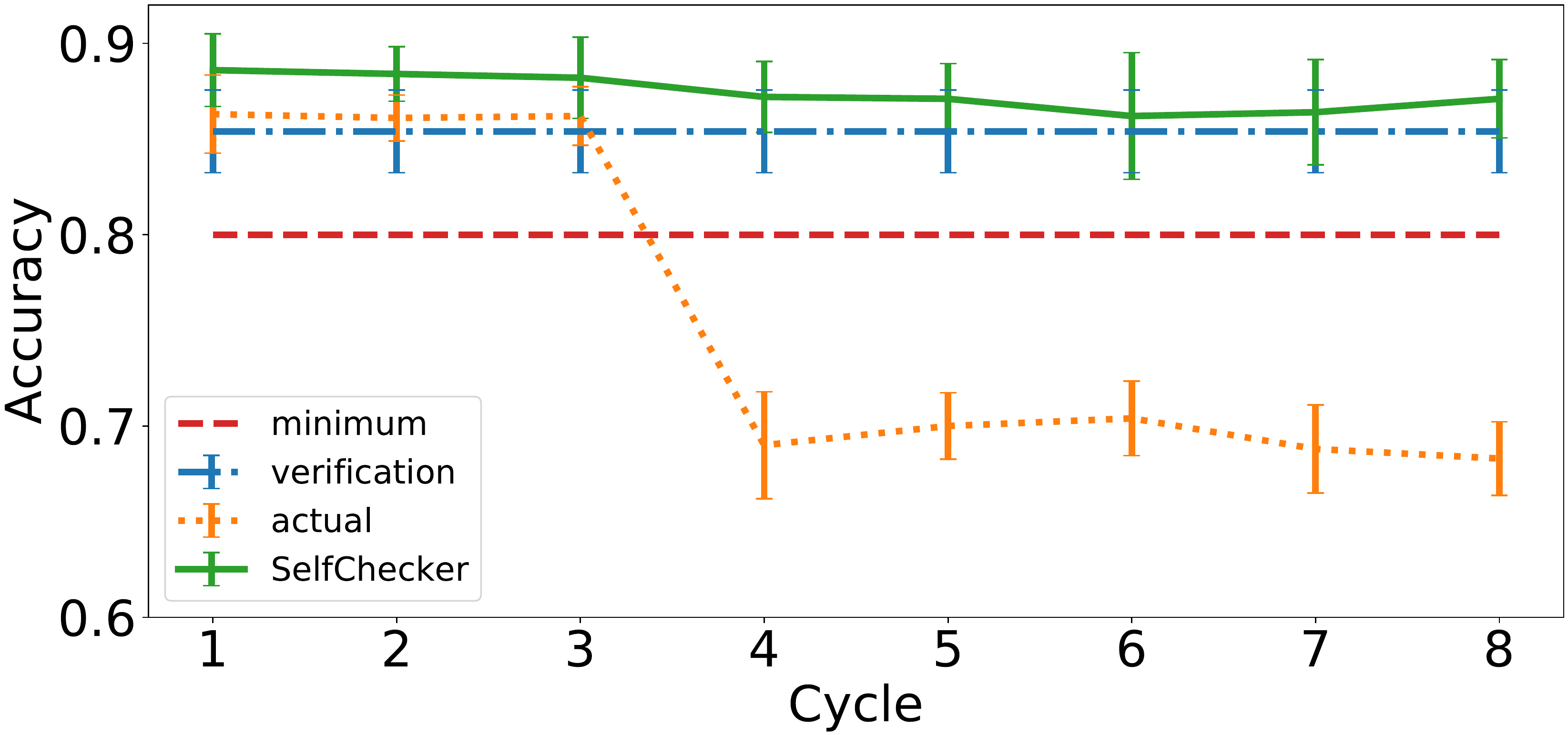}
 	\caption{DAIC results with the SelfChecker pseudo-oracle}
 	\label{fig:loop_sc}
\end{figure}

\begin{table}[t]
\centering
\caption{DAIC detailed results with the SelfChecker pseudo-oracle}
\label{tab:cycleoutput_sc}
	\def\arraystretch{1.2}
\begin{tabular}{c|cccc}
\textbf{} & \multicolumn{1}{c|}{\textbf{verification}} & \multicolumn{1}{c|}{\textbf{actual}} & \multicolumn{1}{c|}{\textbf{predicted acc.}} & \textbf{estimated acc.}\\ 
\textbf{cycle} & \multicolumn{1}{c|}{\textbf{accuracy}} & \multicolumn{1}{c|}{\textbf{accuracy}} & \multicolumn{1}{c|}{\textbf{(SelfChecker)}} & \textbf{(\offlineassessment{})}\\ \hline
1              & \multicolumn{1}{c|}{0.861}                               & \multicolumn{1}{c|}{0.859}                         & \multicolumn{1}{c|}{0.881}                 & \textit{untriggered}               \\ \hline
2              & \multicolumn{1}{c|}{0.861}                               & \multicolumn{1}{c|}{0.863}                         & \multicolumn{1}{c|}{0.883}                 & \textit{untriggered}               \\ \hline
3              & \multicolumn{1}{c|}{0.861}                               & \multicolumn{1}{c|}{0.860}                          & \multicolumn{1}{c|}{0.879}                 & \textit{untriggered}              \\ \hline
4              & \multicolumn{1}{c|}{0.861}                               & \multicolumn{1}{c|}{0.700}                           & \multicolumn{1}{c|}{0.868}                 & \textit{untriggered}             \\ \hline
5              & \multicolumn{1}{c|}{0.861}                               & \multicolumn{1}{c|}{0.698}                         & \multicolumn{1}{c|}{0.861}                 & \textit{untriggered}                \\ \hline
6              & \multicolumn{1}{c|}{0.861}                               & \multicolumn{1}{c|}{0.707}                         & \multicolumn{1}{c|}{0.866}                 & \textit{untriggered}               \\ \hline
7              & \multicolumn{1}{c|}{0.861}                               & \multicolumn{1}{c|}{0.695}                         & \multicolumn{1}{c|}{0.864}                 & \textit{untriggered}            \\ \hline
8              & \multicolumn{1}{c|}{0.861}                               & \multicolumn{1}{c|}{0.692}                         & \multicolumn{1}{c|}{0.867}                 & \textit{untriggered}            \\ \hline
\end{tabular}
\end{table}

\subsection{Results}
DAIC is experimented by running eight cycles, with five repetitions.
The pseudo-oracle assessment is executed at each iteration. 

In the experiments, the triggering condition for the sampling-based assessment is: 

$\{$\textit{predicted accuracy} $<$ (\textit{verification accuracy} - $0.05$)$\}$ OR 

$\{$\textit{predicted accuracy} $<$ \textit{minimum accuracy}$\}$

\noindent that is, the \textit{offline} assessment is triggered when the difference between the accuracy estimated \textit{online} (\textit{predicted accuracy}) and the accuracy estimated prior to release (\textit{verification accuracy}) drops below a given threshold (here set to $0.05$), or when the \textit{predicted accuracy} falls below a \textit{minimum accuracy} required for the DNN (set to $0.80$ in the experiments). 

When \offlineassessment{} is triggered, a set of new manually labeled samples are sent to the Data Preprocessing phase, where they are integrated into the training and verification sets. The proportion between new and old samples in the training dataset may be varied according to the accuracy estimates of last cycle(s). By default, both new and old samples are considered.

\begin{figure}[t]
	\centering
 	\includegraphics[width=0.99\columnwidth]{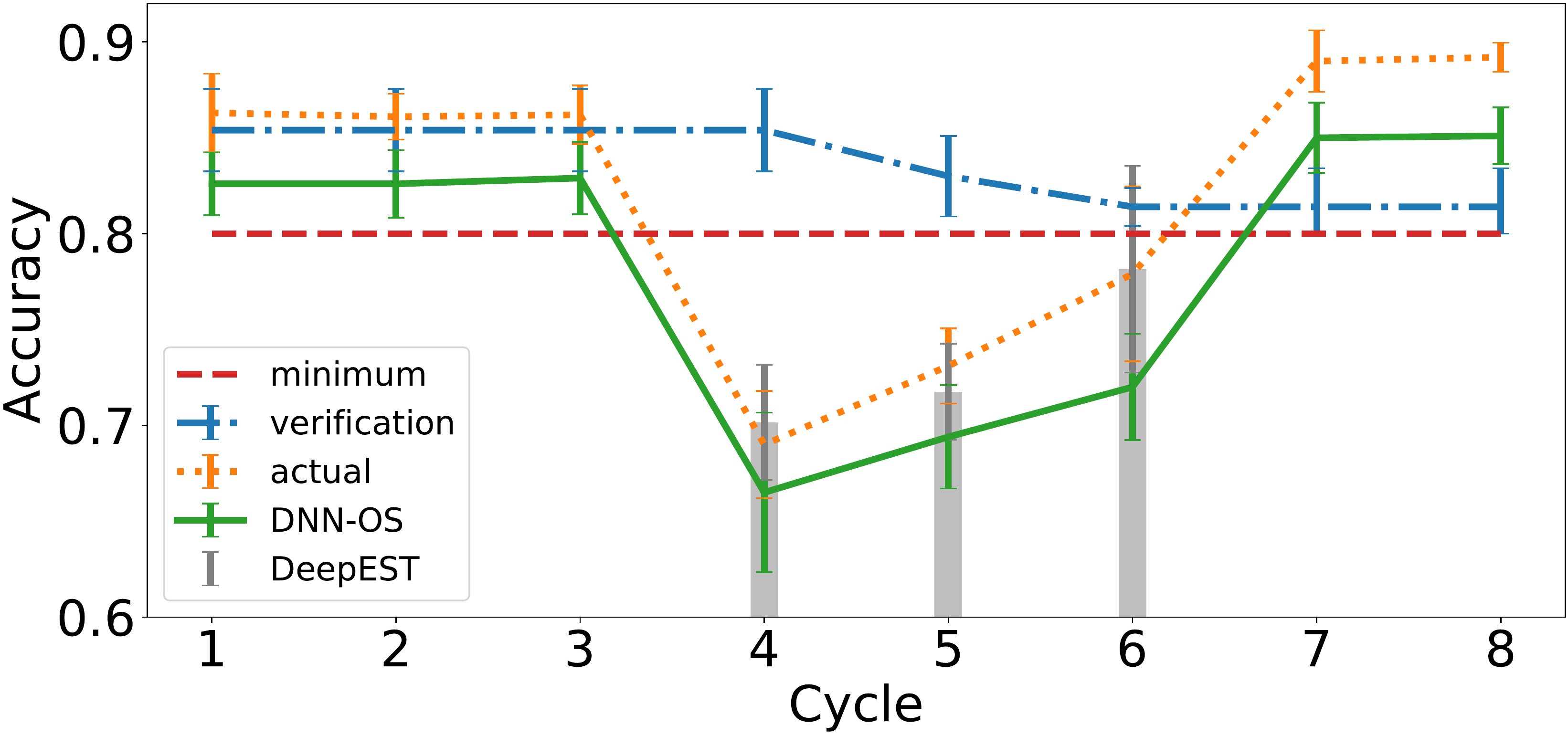}
 	\caption{DAIC results with the \onlineassessment{} pseudo-oracle}
 	\label{fig:loop_icos}
\end{figure}

Figures \ref{fig:loop_sc} and \ref{fig:loop_icos} show the average results and the confidence intervals over 5 repetitions of 8 DAIC iterations. 
Tables \ref{tab:cycleoutput_sc} and \ref{tab:cycleoutput_icos} provide the details for each cycle. 
The first three cycles represent the nominal conditions, namely when the training and validation set faithfully represent the operational dataset. As expected, the operational accuracy computed with SelfChecker (Figure \ref{fig:loop_sc}, and first three rows of Table \ref{tab:cycleoutput_sc}) and with \onlineassessment{} (Figure \ref{fig:loop_icos}, and first three rows of Table \ref{tab:cycleoutput_icos}) does not trigger sampling in the first three cycles.

\begin{table}[t]
\centering
\vspace{1.25pt}
\caption{DAIC detailed results with the \onlineassessment{} pseudo-oracle}
\label{tab:cycleoutput_icos}
	\def\arraystretch{1.2}
\begin{tabular}{c|c|c|c|c}
\textbf{} & \textbf{verification} & \textbf{actual} & \textbf{predicted acc.} & \textbf{estimated acc.}  \\ 
\textbf{cycle} & \textbf{accuracy} & \textbf{accuracy} & \textbf{(\onlineassessment{})} & \textbf{(\offlineassessment{})}  \\ \hline
1              & 0.861                           & 0.859                     & 0.833                   & {\textit{untriggered}}   \\ \hline
2              & 0.861                           & 0.863                     & 0.828                   & {\textit{untriggered}}   \\ \hline
3              & 0.861                           & 0.860                      & 0.825                   & {\textit{untriggered}}   \\ \hline
4              & 0.861                           & 0.700                       & 0.678                   & 0.713        \\ \hline
5              & 0.833                           & 0.726                     & 0.693                   & 0.719          \\ \hline
6              & 0.810                            & 0.781                     & 0.725                   & 0.784         \\ \hline
7              & 0.818                           & 0.888                     & 0.850                    & {\textit{untriggered}}  \\ \hline
8              & 0.818                           & 0.894                     & 0.851                   & {\textit{untriggered}}   \\ \hline
\end{tabular}
\end{table}

Starting from the fourth cycle, a label shift is simulated by switching labels 2 and 7 in operational data. 
As SelfChecker relies only on model and data information, it is unable to detect failures, and its accuracy estimate diverges from the actual one without triggering the assessment via sampling for the remaining cycles (Figure \ref{fig:loop_sc}, and cycle 4 in Table \ref{tab:cycleoutput_sc}).

Thanks to domain invariants, \onlineassessment{} is able to correctly detect failures, providing an accurate estimate of the operational accuracy. 
Indeed, when the accuracy drops (Figure \ref{fig:loop_icos}), it triggers the assessment via sampling (cycle 4 in Table \ref{tab:cycleoutput_icos}). 
\offlineassessment{} estimate confirms that the operational dataset is starting to diverge from the one observed in the previous cycles. Thus, new samples are inserted into the training dataset, and the model is re-trained.
During cycles 5 and 6, the accuracy drops are still correctly detected by \onlineassessment{}, and further improvement actions are performed with the new samples provided by \offlineassessment{}. 

During the 7$^{th}$ cycle the model is trained only with the $1,500$ examples collected in cycles 4, 5, and 6 achieving a high actual accuracy. In both cycles 7 and 8 the accuracy is correctly estimated by \onlineassessment{} as greater than 0.8, avoiding the triggering of \offlineassessment{}.

DNN-OS exhibits a single unnecessary trigger in cycle 6 (repetition 2), where it does not catch that the actual accuracy was already higher than the minimum.

\section{Future Plans}
\label{sec:future}

Life cycles for DNN-based systems adopted in continuous delivery contexts are iterative by nature. The proposed integration of pseudo-oracles and sampling techniques supports both the assessment and the improvement of the DNN accuracy. It helps engineers to leverage collected features in the operational environment to more faithfully evaluate and then specialize the DNN performing the task they need in the way they need.

We plan to refine DAIC defining more sophisticated strategies for the automatic improvement of the DNN in the loop. Techniques like \offlineassessment{} can spot a high number of failing examples, which, along with operational features, can help improve the performance of DNN also in corner cases.
An advancement is to integrate the automatic improvement both at the experimental and deployment stage. As shown in the preliminary results, it is rarely required to change a well-performing model in case of unexpected phenomena like label shift. Often, additional training or training from scratch, by incorporating the operational examples in the training set, may suffice to improve the operational accuracy. For this reason, in line with MLOps perspectives, strategies for the online auto-improvement of DNN can be based on the “probabilistic” output of the pseudo-oracles. Moreover, by automating data preprocessing, the offline re-training step can be run without human intervention.

To a second extent, we plan to apply inferential engines on operational features to automatically extract operational constraints aiming to improve pseudo-oracle effectiveness in estimating the accuracy during the operation.
A recent work from Google stresses the importance of incorporating domain knowledge as a set of rules to improve ML components accuracy \cite{DeepCTL}. %

The iterative assessment and the improvement of the accuracy of the DNN can be of interest beyond the experimented image classification domain. We plan to apply DAIC in industry-relevant domains like Autonomous Driving:
for instance, to the throttle/braking/steering angle prediction, 
which are regression problems. %

\section{Conclusions}
\label{sec:conclusions}
Preliminary results confirm that the accuracy computed before the release can be very different from the one achieved in operation by the DNN in presence of unexpected phenomena like label shift. 
However, the accuracy predicted by \onlineassessment{} follows the actual accuracy with the operational data thanks the \textit{domain} invariants, triggering the assessment via sampling only when required.

The estimates provided by \offlineassessment{} can be used to faithfully evaluate the accuracy provided in operation.
The experimental results also showed that the performance of the DNN in operation can be sensibly increased thanks to the availability of the new labeled examples.

\section*{Acknowledgment}
This project has received funding from the European Union’s Horizon 2020 research and innovation programme under the Marie Sk{\l}odowska-Curie grant agreement No 871342 “uDEVOPS”. It is also supported by the DIETI COSMIC project. 

\bibliographystyle{IEEEtran}
\IEEEtriggeratref{24}

\end{document}